%% file: main.tex
\pdfoutput=1

\documentclass[11pt]{article}

\usepackage[]{EMNLP2023}

\usepackage{times}
\usepackage{latexsym}

\usepackage[T1]{fontenc}

\usepackage[utf8]{inputenc}

\usepackage{microtype}

\usepackage{inconsolata}

\usepackage{mathtools}
\usepackage{adjustbox}
\usepackage{multirow}
\usepackage{url}
\usepackage{latexsym}
\usepackage{algorithm}
\usepackage{algorithmic}
\usepackage{makecell}
\usepackage{booktabs}
\usepackage{listings}
\usepackage{csquotes}
\usepackage{amsthm}
\usepackage{bm}
\usepackage{xspace}
\usepackage{cleveref}
\crefformat{section}{\S#2#1#3}
\crefformat{subsection}{\S#2#1#3}
\usepackage{enumitem}
\usepackage{caption}
\usepackage{subcaption}
\usepackage{microtype}
\usepackage{soul}
\usepackage{amsfonts}

\usepackage{xcolor,colortbl}
\input{math_commands}

\usepackage{amssymb}
\usepackage{amsmath,array}
\crefformat{section}{\S#2#1#3}
\crefformat{subsection}{\S#2#1#3}

\usepackage{arydshln}
\usepackage[english]{babel}
\newtheorem{definition}{Definition}

\newcommand\multiwozentr{\textsc{MultiWoz-Entr}\xspace}
\newcommand\multiwoz{\textsc{MultiWoz}\xspace}
\newcommand\lee{\textsc{Le}\xspace}
\newcommand\augpt{\textsc{AuGPT}\xspace}
\newcommand\mintl{\textsc{MinTL}\xspace}
\newcommand\hdsa{\textsc{Hdsa}\xspace}
\newcommand\marco{\textsc{Marco}\xspace}
\newcommand\sect[1]{\S\ref{#1}}

\definecolor{color1}{rgb}{0.82421875, 0.8671875, 0.859375}
\definecolor{color2}{rgb}{0.5703125, 0.66015625, 0.73828125}
\definecolor{color3}{rgb}{0.75, 0.84375, 0.75}
\definecolor{color4}{rgb}{0.8515625 , 0.90625, 0.984375} 
\definecolor{color5}{rgb}{0.83203125, 0.90625, 0.828125} 
\definecolor{color6}{rgb}{0.921875 , 0.69921875, 0.5625} 
\definecolor{color61}{rgb}{0.98046875, 0.7734375 , 0.52734375} 
\definecolor{color7}{rgb}{0.9609375 , 0.7734375, 0.9140625} 
\definecolor{color8}{rgb}{0.99609375, 0.5703125 , 0.5703125} 
\definecolor{color9}{rgb}{0.99609375, 0.703125  , 0.703125} 
\definecolor{color10}{rgb}{0.99609375, 0.8828125 , 0.46484375} 

\definecolor{Gray}{gray}{0.90}
\definecolor{cid}{HTML}{dae8f5}
\definecolor{ccon}{HTML}{fee9d4}
\definecolor{gred}{HTML}{cc0200}
\definecolor{ggreen}{HTML}{4C9F26}
\newcommand\blue{\cellcolor{cid}}
\newcommand\se{\cellcolor{ccon}}

\title{Lexical Entrainment for Conversational Systems}

\author{Zhengxiang Shi$^{\dagger}$ ~ Procheta Sen$^{\ddagger}$ ~ Aldo Lipani$^{\dagger}$\\
        $^{\dagger}$ University College London, United Kingdom \\ 
        $^{\ddagger}$ University of Liverpool, United Kingdom \\ 
        \texttt{\{zhengxiang.shi.19,aldo.lipani\}@ucl.ac.uk} \\
        \texttt{procheta.sen@liverpool.ac.uk}
}

\begin{document}
\maketitle
\begin{abstract}
\input{paper/0_abstract.tex}
\end{abstract}
\input{paper/1_introduction}
\input{paper/2_lexical_entrainment}

\input{paper/3_model_inspection.tex}
\input{paper/4_dataset}
\input{paper/6_method.tex}
\input{paper/7_experiments.tex}

\input{paper/8_related_work}

\input{paper/9_conclusion.tex}


\section*{Limitations}
\input{paper/9_limitation}

\bibliography{main}
\bibliographystyle{acl_natbib}

\clearpage
\appendix
\label{sec:appendix}
\input{paper/10_appendix}

\end{document}

%% file: math_commands.tex
\usepackage{amsmath,amsfonts,bm}

\def\eqref#1{equation~\ref{#1}}

\def\1{\bm{1}}

\DeclareMathAlphabet{\mathsfit}{\encodingdefault}{\sfdefault}{m}{sl}
\SetMathAlphabet{\mathsfit}{bold}{\encodingdefault}{\sfdefault}{bx}{n}



%% file: paper/0_abstract.tex
Conversational agents have become ubiquitous in assisting with daily tasks, and are expected to possess human-like features. 
One such feature is \textit{lexical entrainment} (\lee), a phenomenon in which speakers in human-human conversations tend to naturally and subconsciously align their lexical choices with those of their interlocutors, leading to more successful and engaging conversations. 
As an example, if a digital assistant replies ``Your \underline{appointment} for Jinling Noodle \underline{Pub} is at 7 pm'' to the question ``When is my \underline{reservation} for Jinling Noodle \underline{Bar} today?'', it may feel as though the assistant is trying to correct the speaker, whereas a response of 
``Your \underline{reservation} for Jinling Noodle \underline{Bar} is at 7 pm'' would likely be perceived as more positive. This highlights the importance of \lee in establishing a shared terminology for maximum clarity and reducing ambiguity in conversations. 
However, we demonstrate in this work that current response generation models do not adequately address this crucial human-like phenomenon. 
To address this, we propose a new dataset, named \multiwozentr, and a measure for \lee for conversational systems. 
Additionally, we suggest a way to explicitly integrate \lee into conversational systems with two new tasks, a \lee extraction task and a \lee generation task. We also present two baseline approaches for the \lee extraction task, which aim to detect \lee expressions from dialogue contexts.\footnote{The dataset is available at 
\url{https://github.com/ZhengxiangShi/LexicalEntrainment}.}

%% file: paper/1_introduction.tex
\section{Introduction}

\begin{figure}[!t]
  \centering
  \includegraphics[width=\columnwidth]{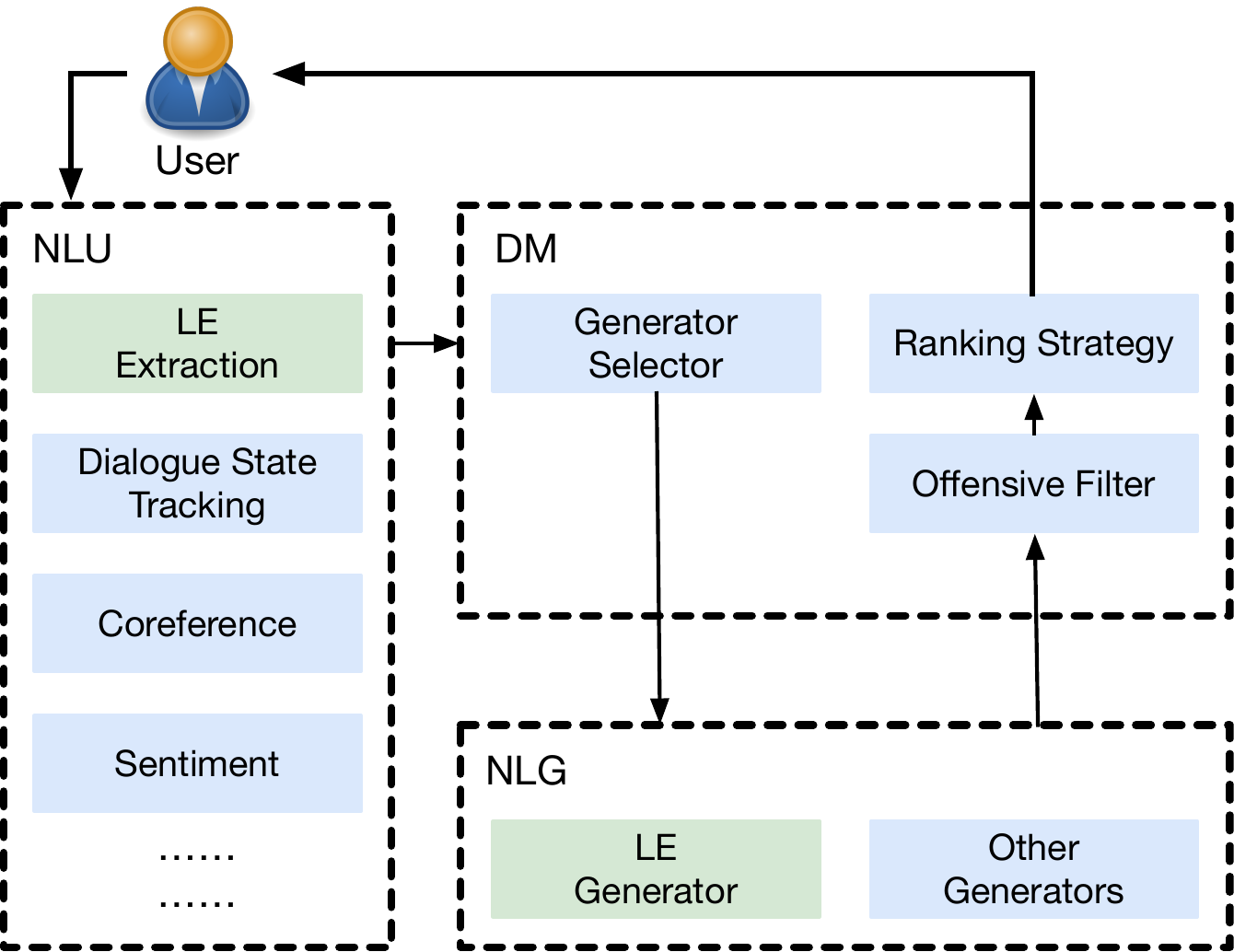}
  \caption{Illustration of a (simplified) pipeline architecture of conversational systems, which decomposes into \textsc{NLU}, \textsc{DM}, and \textsc{NLG} modules. \lee expressions extracted by the \textsc{NLU} module can be leveraged by the NLG module to generate more engaging and human-like responses.} 
  \label{fig:flow_chart}
\end{figure}

There is potential for enormous variability in people's lexical choices in dialogues, whether they are communicating with human or machine partners \cite{brennan1996lexical}. One question that has attracted some attention from the linguistics community has been: \textit{How do people deal with such high lexical variability when they talk to each other}? 
In the referential communication studies \cite{brennan1996lexical1,brennan1996lexical}, the likelihood that people in one conversation would choose the same terms to refer to the same common object as people in another conversation was only 10\%. 
Despite the high variability across conversations, it is relatively low within a conversation \cite{garrod1987saying,brennan1996lexical}. When people engage in a conversation, they naturally adapt their way of speaking to align with their conversational partner. 
For instance, they tend to refer to something based on how their conversational partners refer to it, using the same terms when discussing the same object repeatedly and negotiating a common description, particularly for items that may be unfamiliar to them \cite{brennan1996lexical1}. This linguistic phenomenon is known as \textit{lexical entrainment} (\lee) \cite{garrod1987saying,brennan1996lexical}.

\lee plays a key role in the success and naturalness of interactions in conversations. \citet{reitter2007predicting} demonstrated that lexical repetition is a reliable predictor of task success given the first five minutes in task-oriented dialogues. 
The degree of entrainment with respect to the most frequent words is also a distinguishing factor between dialogues rated as most natural and those rated as less natural, and is strongly correlated with task success and engaged and coordinated turn-taking behaviour \cite{nenkova2008high}.
Additionally, \lee is associated with a broad range of positive social behaviours and outcomes \cite{lopes2015rule, nasir2019modeling}, such as building effective dialogues \cite{porzel2006entrainment} and engaging students in tutorial dialogues \cite{ward2007measuring}. 
In conversational systems, an explicit mechanism of lexical choice (e.g., by \lee) is needed to ensure meaningful results \cite{pickering2004toward}, thus making conversations more engaging and successful.

Despite its importance, \lee has not been systematically studied for conversational systems, and this interesting fact of language-based communication has been overlooked. 
There is a scarcity of datasets specifically designed to study \lee, and it is not explicitly modelled in any state-of-the-art conversational systems. 
The only existing dataset for lexical entrainment, presented by \citet{duvsek2016context1}, consists of $1\,859$ pairs of user utterances, system responses, and dialogue acts.
However, it is limited to the context of a single preceding user utterance and constrained to the public transportation domain. Most importantly, this dataset lacks explicit annotations of \lee, making it difficult to be used to explicitly model \lee in conversational systems.


To this end, we propose a new dataset named \multiwozentr with annotated \lee information, comprehensive dialogue context, and detailed statistical analysis. 
The dataset is built on the top of the \multiwoz \cite{eric2019multiwoz}, a task-oriented dialogue dataset with 7 distinct domains. 
\multiwoz is selected as a starting point for two reasons: (1) it is a task-oriented dataset, which tends to contain more \lee expressions than other kinds of dialogue datasets \cite{reitter2006computational}, such as open-domain datasets \cite[e.g.,][]{li2017dailydialog, wu2017sequential}; and (2) it is collected based on human-to-human conversations.  

Additionally, we present a workflow of a conversational system integrated with \lee modules, as illustrated in Figure \ref{fig:flow_chart}. 
A standard pipeline architecture for conversational systems generally decomposes into three modules, a Natural Language Understanding (\textsc{NLU}) module, a Dialogue Management (\textsc{DM}) module, and a Natural Language Generation (\textsc{NLG}) module. 
Each of these modules consists of several sub-modules that can be modelled independently. 
The integration of \lee could be accomplished through two additional sub-modules (tasks), a \lee extraction sub-module and a \lee generator sub-module. 
The objective of the \lee extraction sub-module is to identify lexical candidates and extract \lee expressions given the dialogue context. 
The objective of the generator module is to make use of these extracted lexical candidates to generate more human-like responses. Additionally, we provide two baseline models for the \lee extraction sub-module.

In summary, the main contributions of this paper are as follows:
\begin{itemize}
\item We formalize a precise and practical definition of \lee (\cref{sec:definition}), as well as a new \lee measure to evaluate the natural degree of \lee in human-to-human conversations (\cref{sec:new_measure});
\item We highlight the importance of \lee and provide an analysis of state-of-the-art response generation models, pointing out issues caused by the neglect of \lee (\cref{sec:model_inspection});
\item We propose a \lee dataset, \multiwozentr, specifically designed for studying \lee, and provide detailed annotations (\cref{sec:dataset}) and statistical analysis (\cref{sec:dataset_statistics});
\item We present a methodology for integrating \lee into conversational systems through two novel sub-modules (tasks). Specifically, we provide two baseline models for the \lee extraction sub-module (\cref{sec:method}), providing valuable insights into the incorporation of \lee into conversational systems. This approach lays the groundwork for the development of a \lee generator sub-module, which is left for future research (\cref{sec:main_experiment}).
\end{itemize}

%% file: paper/2_lexical_entrainment.tex

\section{Lexical entrainment (\lee)}
\input{table/example}
\subsection{Definition of \lee}
\label{sec:definition}
According to the work of \citet{brennan1996lexical}, \lee is defined as ``\textit{[w]hen two people repeatedly discuss the same object, they come to use the same terms.}'' 
\citeauthor{brennan1996lexical} also provides several examples of entrained expressions, which are noun phrases and also referred to as \textit{referring expressions}. 
However, this definition and the examples provided are not specific, leaving room for multiple interpretations.
This has led to several works in the field of computer science with conflicting interpretations. 
For instance, \citet{nenkova2008high,duvsek2016context1,hu2016entrainment,duplessis2017automatic,duplessis2021towards} 
have expanded upon \citeauthor{brennan1996lexical}'s definition of \lee to include other parts of speech beyond noun phrases,
such as modal particles (e.g., ``well'') and verb phrases (e.g., ``work for''). 
This work aims to eliminate ambiguity and formalize a precise and operational definition of \lee, which closely aligns with the original definition proposed by \citet{brennan1996lexical}.
Accordingly, the following definition is proposed:

\begin{definition}[\textbf{Lexical Entrainment}] 
Lexical entrainment refers to the natural phenomenon observed in conversations involving two or more speakers, in which \textit{equivalent} referring expressions are utilised to discuss the same object. A referring expression is a noun phrase used to denote an object, which can refer to both concrete and abstract entities, and may be singular, plural, or collective.
\end{definition}
%
In the original work by \citet{brennan1996lexical1}, it was suggested that two referring expressions can be considered equivalent only if they share all the same content words. For instance, expressions such as ``the red dog with its mouth open'' and ``the dog with its mouth open'' would not be regarded as equivalent.
However, we find this definition to be excessively stringent, and thus, we propose a more relaxed criterion that allows for different modifiers in the referring expressions.
As an example, consider the dialogue in Table~\ref{tbl:example}, where ``the reference number for that reservation'' in the $17^{\texttt{th}}$ utterance and ``your reference number'' in the $18^{\texttt{th}}$ utterance would not satisfy \citeauthor{brennan1996lexical1}'s criterion.
Despite this, the agent's choice to use the expression ``reference number'' implies a lexical decision to entrain the user, given that the agent could have used alternative expressions like ``confirmation number''. 
Therefore, we consider the two expressions to be equivalent after removing the modifier ``for that reservation''. 
Another example of equivalent referring expressions is ``red taxi'' and ``blue taxi'', which can be considered equivalent after disregarding the adjectives ``red'' and ``blue'', given that the agent could have used alternative expressions like ``blue cab''.
The concept of equivalent referring expressions is defined as follows:

\begin{definition}[\textbf{Equivalent Referring Expressions}]
Two referring expressions are deemed equivalent if, without their modifiers, their head nouns are the same, regardless of their noun forms.
\end{definition}


In accordance with the terminology established by \citet{duplessis2017automatic}, referring expressions that are equivalent in a conversation are referred to as \textbf{a \lee expression} once they have been \textbf{established}. 
Each referring expression is then considered an instance of this \lee expression.
For example, ``table'' and ``tables'' in Table~\ref{tbl:example} belong to the same \lee expression, and `table'' and ``tables'' themselves are considered two instances of this \lee expression.
\textbf{An instance of \lee expression} can either be free or constrained given a dialogue. 
\textbf{A free instance} is an instance of a \lee expression that appears without being a subexpression of a larger \lee expression, 
while \textbf{a constrained instance} is an instance of a \lee expression that appears as a subexpression of a larger \lee expression. 
For instance, in Table \ref{tbl:example}, ``italian restaurant'' in the $1^{\texttt{st}}$ and $2^{\texttt{nd}}$ utterances, as well as ``restaurant'' in the $3^{\texttt{rd}}$ utterance, are all free instances, 
while "restaurant" in the $1^{\texttt{st}}$ and $2^{\texttt{nd}}$ utterances are constrained instances since they are sub-expressions of ``italian restaurant''. 
Equivalent referring expressions are \textbf{established} if and only if the two following requirements are met: 
(i) they have been produced by at least two speakers, either in free or constrained form, and 
(ii) at least one of the instances is in free form. 
For example, ``italian restaurant'' is established in the $2^{\texttt{nd}}$ utterance and ``price range'' is established in the $11^{\texttt{th}}$ utterance.
Finally, the \textbf{initiator of a \lee expression} refers to the speaker who first produces an instance of the \lee expression, either in a free or constrained form. 

\subsection{The proposed measure of \lee}
\label{sec:new_measure}

We propose a new measure, referred to as the degree of \lee, to evaluate the frequency of \lee expressions in dialogues after an agreement has been reached (i.e., \lee expressions are established). Given a dialogue with $n$ utterances from the speaker $s$, this measure is computed as follows:
\begin{equation}
    \textsc{Entr}_s = \frac{1}{n} \sum^{n}_{j=1} E_{s,j},
\end{equation}
where $E_{s,j}$ is the number of instances of an \lee expression in the $j^{\texttt{th}}$ utterance from the speaker $s$ of established \lee expression.

It is worth noting that a higher value of $\textsc{Entr}_s$ does not necessarily imply higher quality of the conversation, as conversational systems should aim to find a balance between self-consistency and \lee rather than simply maximising the degree of \lee. Nevertheless, the degree of \lee can be used to understand the natural frequency of \lee in human-to-human conversations and what might be pertinent in the design of machines.


%
%


%

%


%% file: table/example.tex
\newcommand\myfontsizeeee{\fontsize{5.8pt}{6.5pt}\selectfont}
\begin{table*}[hbt!]
    \centering
    \myfontsizeeee
    \setlength{\tabcolsep}{5.75pt}
    \renewcommand{\arraystretch}{1.2}
    \caption{An example from \multiwozentr. The sequence of words with the same colour represents instances of the same \lee expressions. 
    The sequence of words with the underline stands for the \lee expressions entrained in that utterance, meaning that this \lee expression has already been established by the current utterance. 
    $E_{s,j}$ indicates the number of instances of established \lee expressions in the $j^{\texttt{th}}$ turn from the speaker $s$.}
    \begin{adjustbox}{max width=\textwidth}
    \begin{tabular}{p{0.02\linewidth}|p{0.045\linewidth}|p{0.83\linewidth}|p{0.02\linewidth}}
      \toprule
      {\scriptsize \bf No.} & {\scriptsize \bf Speaker} & {\scriptsize \bf Utterance} & \bf $\bf E_{s,j}$\\
      \midrule
        1   & User                        & Im looking for an \colorbox{color2}{italian \colorbox{color2!50}{restaurant}}. I also need it to be near the \colorbox{color61}{\colorbox{color61!50}{center} of town}. & 0   \\
        2   & Agent                       & There are 9 \colorbox{color2}{\underline{italian \colorbox{color2!50}{restaurants}}} in the \colorbox{color61}{\underline{\colorbox{color61!50}{center} of town}}. Do you have a particular \colorbox{color4}{price range} in mind?  & 2   \\
        3   & User                        & I need a cheap \colorbox{color2!50}{\underline{restaurant}}.  & 1 \\
        4   & Agent                       & There is Pizza Hut City Centre, \colorbox{color10}{Ask}, and Zizzi Cambridge, which would you like to book?  & 0 \\
        5   & User                        & Can you book a \colorbox{color9}{table} for 7 at \colorbox{color10}{\underline{Ask}}? We're looking at 14:00 on Saturday.  & 1 \\
        6   & Agent                       & Sorry, there are no \colorbox{color9}{\underline{tables}} available at \colorbox{color10}{\underline{Ask}} at that time. Would you like to book another day or time, or perhaps another \colorbox{color2!50}{\underline{restaurant}}?  & 3 \\
        7   & User                        & Can you try 13:00, then?    & 0 \\
        8   & Agent                       & 13:00 it is. Your \colorbox{color9}{\underline{table}} at \colorbox{color10}{\underline{Ask}} will be held for 15 minutes, and your \colorbox{brown!50}{reference number} is L2KZYL26. Can I help you with anything else?    & 2  \\
        9   & User                        & I'm also looking for a \colorbox{color5}{hotel}.                                                         & 0        \\
        10   & Agent                       & We have 33 different \colorbox{color5}{\underline{hotels}}.  Did you have a \colorbox{color4}{price range} or star rating in mind?  & 1 \\
        11  & User                        & Something in the cheap \colorbox{color4}{\underline{price range}} including free parking.                    & 1   \\
        12  & Agent                       & What area would you prefer to stay in? This will help narrow down your options.  & 0 \\
        13  & User                        & I would like to book a \colorbox{color5}{\underline{hotel}} close to our \colorbox{color2!50}{\underline{restaurant}} here in the \colorbox{color61!50}{\underline{centre}}, it will also need to be cheap, and have free parking.  & 2 \\
        14  & Agent                       & There are two \colorbox{color5}{\underline{hotels}} available. Would you like me to book one for you?  & 1  \\
        15  & User                        & Yes please. The our group of 7 will all be staying there for 3 nights starting\colorbox{white}{Saturday.}      & 0 \\
        16  & Agent                       & I've booked you a 3 day stay at the Alexander Bed and Breakfast for 7 people. Do you need anything\colorbox{white}{else?}         & 0  \\
        17  & User                        & Could I have the \colorbox{brown!50}{\underline{reference number}} for that reservation please?      & 1  \\
        18  & Agent                       & Yes, your \colorbox{brown!50}{\underline{reference number}} is 1JBO1R08.      & 1 \\
        19  & User                        & Thank you so much for all of your help. That is all I needed for now. Have a nice \colorbox{color6}{day}.     & 0 \\
        20  & Agent                       & Thank you for choosing us to assist you on your travels. Have a good \colorbox{color6}{\underline{day}}.        & 1  \\ 
      \bottomrule
    \end{tabular}
    \end{adjustbox}
    \label{tbl:example}
\end{table*}

%% file: paper/3_model_inspection.tex
%

\input{table/case_study}

\section{Limitations of current approaches}
\label{sec:model_inspection}
Current conversational systems do not sufficiently address the phenomenon of \lee, leading to inconsistencies between generated responses and actual human responses. This can result in sub-optimal generated responses, which are not captured by current evaluation metrics, such as \textsc{Inform Rate}, \textsc{Success Rate}, and \textsc{Bleu Score}.
To demonstrate the discrepancies between generated responses and ground-truth responses, we evaluate four state-of-the-art response generation models \cite{nekvinda-dusek-2021-shades} in both end-to-end fashions, that is, using only the dialogue context as input to generate responses, and policy optimization fashion, that is using the ground-truth dialogue states to generate responses. For end-to-end models, \mintl \cite{lin2020mintl} and \augpt \cite{kulhanek-etal-2021-augpt} are evaluated. For policy optimization models, \hdsa \cite{chen-etal-2019-semantically} and \marco \cite{wang-etal-2020-multi-domain} are evaluated. 
In the following, we present two examples of generated responses that demonstrate the consequences of neglecting \lee.

%
In the first case study (see Table \ref{table:case1}), the user requests a ``confirmation number'', and the ground-truth (human-like) agent response entrains this expression by repeating ``confirmation number''. 
However, all four generated responses utilise ``reference number'' instead. 
This is due to the fact that ``reference number'' appears more frequently in the training data or is used in their generation templates.
In this scenario, using a ``confirmation number'' would be more appropriate and engaging for the user.
In the second case study, the user requests a ``cab'', which is also entrained by the ground-truth agent response.
However, \marco replies to the user with the ``taxi''.
Many similar situations exist where generation models struggle to choose synonyms, such as ``phone number,'' ``contact number,'' and ``telephone number,'' as well as ``good reviews'' and ``good ratings.''

The above-mentioned issues stem from neglecting the phenomenon of \lee, which occurs naturally and unconsciously in human-to-human conversations. We consolidate our observation with a quantifiable analysis in \cref{sec:quantitive_analysis}.

%% file: table/case_study.tex
\newcommand\secondfontsize{\fontsize{7.5pt}{9.0pt}\selectfont}
\begin{table*}[hbt!]
    \centering
    \scriptsize
    \setlength{\tabcolsep}{5.75pt}
    \renewcommand{\arraystretch}{1.2}
    \caption{
    Case study for \lee in existing conversational systems.
    The sequence of words with the underline stands for the \lee expressions entrained in that utterance, meaning that this \lee expression has already been established by the current utterance. 
    In case study 1, when the user asks for a ``\textit{confirmation number}'', response generation models consistently provide a response using the term ``\textit{reference number}'' instead. In case study 2, the user requests the agent to arrange a ``\textit{cab}'' but the response generation model provides a response using the term ``\textit{taxi}''.}
    \begin{adjustbox}{max width=\textwidth}
    \begin{tabular}{p{0.045\linewidth}|p{0.92\linewidth}}
      \toprule
      \bf Model & \bf Utterance (Case Study 1) \\
      \midrule
      \textsc{User}  &  Yes, please book it for 2 people on Saturday for 4 nights. Will you give me the \colorbox{brown!45}{confirmation number} when that is complete? \\
      \textsc{Agent} &  Sure, booking was successful. Your \colorbox{brown!45}{\underline{confirmation number}} is 7MG032V9.  \\
      \midrule
      \augpt &  Booking was successful. \textcolor{brown}{Reference number} is: 00000016. \\
      \mintl &  I have booked you at the Express by Holiday Inn Cambridge for 4 nights starting Saturday. Your \textcolor{brown}{reference number} is D77JCGJ2. Is there anything else I can help? \\
      \hdsa &  Booking was successful. \textcolor{brown}{Reference number} is 7MG032V9. Is there anything else I can help you with? \\
      \marco & I have booked you at the \texttt{[hotel\_name]}. Your \textcolor{brown}{reference number} is 7MG032V9. Is there anything else I can help you with? \\
      \midrule
      \bf Model & \bf Utterance (Case Study 2) \\
      \midrule
      \textsc{User}  &  I need a \colorbox{color5}{cab} too please, arriving by my \colorbox{color4}{booked time}, car type and contact number would be helpful too please. \\
      \textsc{Agent} & You need a \colorbox{color5}{\underline{cab}} to get you to the restaurant before the \colorbox{color4}{\underline{booked time}} and where do you need \colorbox{color5}{\underline{cab}} to pick you up at? \\
      \midrule
      \augpt &  I have booked a blue Skoda for you. The contact number is 85534112262. \\
      \mintl &  I have booked you a BMW. The contact number is 07338019809. Where will you be departing from? \\
      \hdsa  &  Where will you be departing from and where are you going? \\
      \marco &  I would be happy to book a \textcolor{teal}{taxi} for you . where are you departing from and what is your destination? \\
      \bottomrule
    \end{tabular}
    \end{adjustbox}
    \label{table:case1}
\end{table*}

%% file: paper/4_dataset.tex
\section{\multiwozentr}
\label{sec:dataset}
To facilitate \lee research for conversational systems, 
we introduce a new dataset named \multiwozentr based on the \multiwoz 2.1 \cite{eric2019multiwoz}.

\subsection{\multiwoz.} \label{appendix:woz}
\multiwoz 2.1 contains $8\,438$, $1\,000$, and $1\,000$ samples for training, validation, and test sets respectively.
It spans over seven domains, including Taxi, Hotel, Attraction, Train, Restaurant, Police, and Hospital. 
The Police and Hospital domains are not included in the validation and testing sets, but the training set contains 245 and 287 samples for these domains, respectively.
Detailed statistics for the other five domains are summarised in Table \ref{tbl:multiwoz_statistics}. 
\input{table/statistics_multiwoz}

\subsection{\multiwozentr.}
We use the following steps to generate annotations of \lee expressions in \multiwoz.
Please refer to \cref{sec:dataset_statistics} for a detailed statistical analysis of the proposed dataset.


\paragraph{\textbf{Step 1: Preprocessing.}} 
The goal of this step is to ensure that all the \lee expressions can be easily captured from the given dialogues.
In previous works \cite[e.g.,][]{duplessis2017automatic, duplessis2021towards}, 
two text sequences are regarded equal only if they are exactly the same. 
However, even if they contain \lee expressions, two sentences may have minor grammatical differences, such as different forms (e.g., ``table'' and ``tables''), British and American English spelling (e.g., ``centre'' and ``center''), and numerical characters (e.g., ``4 star hotel'' and ``four star hotel'').
To tackle these issues, we first convert all British English to American English using an open-source toolkit, thereby converting ``centre'' to ``center'', for example.
We also standardize numerical characters. 
We then use a stemming algorithm with the NLTK toolkit \cite{bird2004nltk} for each token in a dialogue. 
For example, "italian restaurant" and "italian restaurants" in Table~\ref{tbl:example} will both be converted to "italian restaur", which ensures that they can be recognized in the next step. 
Meanwhile, we replace all punctuation (except in-word punctuation such as "pre-trained") with pseudo-random numbers with $k$ random bits to prevent them from being identified as parts of \lee expressions in the subsequent step. 
In our work, \lee expressions are punctuation-free and not case-sensitive, which are factors that were not considered in previous works \cite[e.g.,][]{duplessis2017automatic, duplessis2021towards}. 

\paragraph{\textbf{Step 2: Maximize recall.}} 
The target of this step is to enhance the recall rate while extracting potential candidates of \lee expressions from the given dialogues.
To achieve this, we utilise an open-source toolkit\footnote{\url{https://github.com/GuillaumeDD/dialign}} \cite{duplessis2021towards}, which is specifically designed for \lee in dyadic conversations. 
The toolkit retrieves all sequences of tokens that share the same surface text form as that used by the two interlocutors.
Nonetheless, this approach may introduce noise by including irrelevant phrases such as ``in the'', ``for 7'', and ``like to book''.

\paragraph{\textbf{Step 3: Maximize precision.}} 
The purpose of this step is to eliminate noise from the pool of candidates for \lee expressions.
We post-process the candidates of \lee expressions by removing stopwords \cite{bird2004nltk} from the beginning or at the end of \lee expression candidates.
For example, ``the postcode'' will be converted to ``postcode'' and ``in the center'' will be converted to ``center''. 
However, stopwords are allowed to appear in the middle of \lee expression candidates, for instance, ``cheap hotel with free park'' will not be post-processed since ``with'' is a stopword in the middle of this expression.
Additionally, we only consider noun phrases as per the definition of \lee, introduced in \cref{sec:definition}.
To identify noun phrases, we manually create dictionaries for verbs and adjectives/adverbs in the \multiwoz dataset, along with a dictionary for undesired phrases such as particles, modal verbs, and prepositions. 
This helps filter out negative examples from the candidate pool. 
However, some special cases require rule-based methods.
For example, some words such as ``help'' and ``booking'' can function as both nouns and verbs, making it difficult to accurately distinguish them.
In these cases, we examine the special cases and apply specific rules.
Since \multiwoz is a task-oriented dataset with a limited lexicon size, our rule-based methods are feasible and appropriate.

\paragraph{\textbf{Step 4: Human validation}} This step aims to ensure the quality of the \multiwozentr dataset by conducting a human validation process.
The authors of the dataset serve as validators to assess the annotations made in the previous steps.
At least two human validators review every annotated \lee expression, and any disagreements among validators are resolved through discussion.
To perform this validation, a batch of 50 dialogues from all $10\,438$ dialogues without replacement is randomly selected, and all \lee expressions within each dialogue are manually annotated by the validators as the ground-truth labels.
For each dialogue, the automatic annotations obtained in the previous three steps are compared with the ground-truth labels to compute the recall and precision rates, and the $F_1$ Score is computed. If the $F_1$ Score is not 100\%, additional rule-based methods are applied in the previous steps. This annotation process is repeated until successive 100\% $F_1$ scores are achieved.
This human validation step took $24$ runs in total.
This validation process is effective because \multiwoz is a task-oriented conversational dataset with a limited range of variations in sentence patterns and lexicons.

%% file: table/statistics_multiwoz.tex
\begin{table}[ht!]
\centering
\footnotesize
\caption{Statistics of MultiWOZ 2.1: the number of dialogues in different domains.}
\begin{adjustbox}{max width=\columnwidth}
\begin{tabular}{lccccc}
\toprule
\bf Domain & \bf Taxi & \bf Hotel & \bf Attraction & \bf Train & \bf Restaurant \\ \midrule
Train  & $1\,655$ & $3\,387$  & $2\,718$       & $3\,117$  & $3\,817$       \\
Valid  & 207  & 416   & 401        & 484   & 438        \\
Test   & 195  & 394   & 396        & 495   & 437        \\ \bottomrule
\end{tabular}
\end{adjustbox}
\label{tbl:multiwoz_statistics}
\end{table}

%% file: paper/6_method.tex
\section{\lee extraction}
\label{sec:method}
As depicted in Figure~\ref{fig:flow_chart}, to incorporate the \lee phenomenon into conversational systems, it is necessary first to identify lexical choices that can potentially be used as \lee expressions through an extraction process.
The resulting candidates can then be fed into \lee generators to produce appropriate responses.
In this section, we introduce a novel task of extracting \lee expressions from the preceding dialogue context and present two baseline models: an end-to-end approach and a pipeline approach.

\subsection{\lee extraction task definition}
\label{sec:task}
In this section, we introduce the task of extracting \lee expressions.
The task involves identifying all instances of \lee expressions established in or before a given utterance, denoted by underlined text in Table~\ref{tbl:example}. 
For example, in Table~\ref{tbl:example}, if we aim to extract \lee expressions for the $2^{\texttt{nd}}$ utterance, 
the model must be able to identify all occurrences of ``italian restaurant'' and ``center of town'' that have already been established at the $2^{\texttt{nd}}$ utterance (thus they are underlined) from the preceding dialogue context.
However, ``price range'' will not be considered until the $11^{\texttt{th}}$ utterance is established. 
For each sample, the model receives the preceding dialogue context as input, and its goal is to identify all instances of established \lee expressions in the target utterance. 

\subsection{End-to-end approach}
\label{sec:end2end_approach}

The end-to-end model comprises three major components: \textit{dialogue context encoder}, \textit{contingent fusion module}, and \textit{linear layer}. 
At first, the model uses the dialogue context encoder \cite{devlin2018bert} to encode the dialogue context $C$ (separated by ``Human'' and ``Agent'' annotations) into a sequence of token embeddings $\{u_j\}_{j=1}^{s}$, where $u_j \in \mathbb{R}^{d_w}$ is the embedding for the $j^{\texttt{th}}$ token and $s$ is the sequence length of the dialogue context.
Next, the fusion module, stacked by $l$ sub-layers where each sub-layer consists of one attention layer and one feed-forward layer, is employed to update the token embeddings if and only if dialogue acts of agent utterances are inputted. For each turn, the dialogue act, whose tokens of slot and its value are concatenated together, is represented as a vector $x \in \mathbb{R}^{d_w}$ via the dialogue context encoder to update token embeddings $\{u_j\}_{j=1}^{s}$ via the attention mechanism \cite{bahdanau2014neural}: $u_j^{l}=\sum_{j} a_j u_j^{l-1}$, where $a_j = \frac{exp(s_j^{l-1})}{\sum_{j=1}^{K}exp(s_j^{l-1})}$ and $s_j^{l-1}$ is the matching score computed based on the query vector $x$ and each context vector $u_j^{l-1}$. 
Finally, each token embedding $u_j$ is then mapped into a scalar score using a trainable linear layer to minimize the cross-entropy loss.

Given the ground-truth token labels for each token in the dialogue context, denoted as $y = \{y_1, y_2, \dots ,y_N\}$, we minimize the cross entropy loss to predict whether the token belongs to the entrainment expression: $\mathcal{L}=- \sum_{j} [\alpha \cdot y_j \log \hat{y_j} + (1 - y_j) \log(1 - \hat{y_j)} ]$, where the hyperparameter $\alpha$ can be increased to improve the recall rate or reduced to increase the precision rate.

\subsection{Pipeline approach}
The pipeline approach for \lee expression extraction involves two distinct steps: \textit{named entity recognition} and \textit{classification}. In the first step, the dialogue context is provided to a named entity recognition model, which extracts all the noun phrases from the context. In the next step, these extracted phrases are concatenated with the dialogue context and fed into a binary classifier model to determine whether each phrase represents a \lee expression. For the named entity recognition model, we employ NLTK \cite{bird2004nltk}, and for the binary classifier model, we use \textsc{Bert} \cite{devlin2018bert} with an additional trainable linear layer. The fusion module described in \cref{sec:end2end_approach} is utilized if system acts are included as input.

%% file: paper/7_experiments.tex
\section{Experiments and Results}
\label{sec:main_experiment}
In this section, we first implement quantitative analysis using our new proposed \lee measure. Then we evaluate two baseline models on the \multiwozentr for the \lee extraction task. Additionally, we study the effect of different domains on the model performance.

\input{table/anaylsis.tex}
\subsection{Quantitative Analysis}
\label{sec:quantitive_analysis}
To evaluate the discrepancy in \lee between generation models and the ground-truth responses, we compute the average value of $\textsc{Entr}_{agent}$ for each response generation model in the test set.
Table~\ref{tbl:analysis} presents the results, which show that the ground-truth agent (Human) tends to entrain the user about 0.8 times on average, while the four response generation models entrain the user with a lower degree of frequency.
Specifically, in Table \ref{tbl:example}, $\textsc{Entr}_{user}$ is 0.6 and $\textsc{Entr}_{agent}$ is 1.1, which is an average value calculated across utterances. 
The t-test results reveal a significant difference in the means of $\textsc{Entr}_{agent}$ between the generation model and ground truth responses, indicating that there is considerable scope for improvement in state-of-the-art response generation models in terms of incorporating \lee.

\begin{figure}[!t]
  \centering
  \includegraphics[width=\columnwidth]{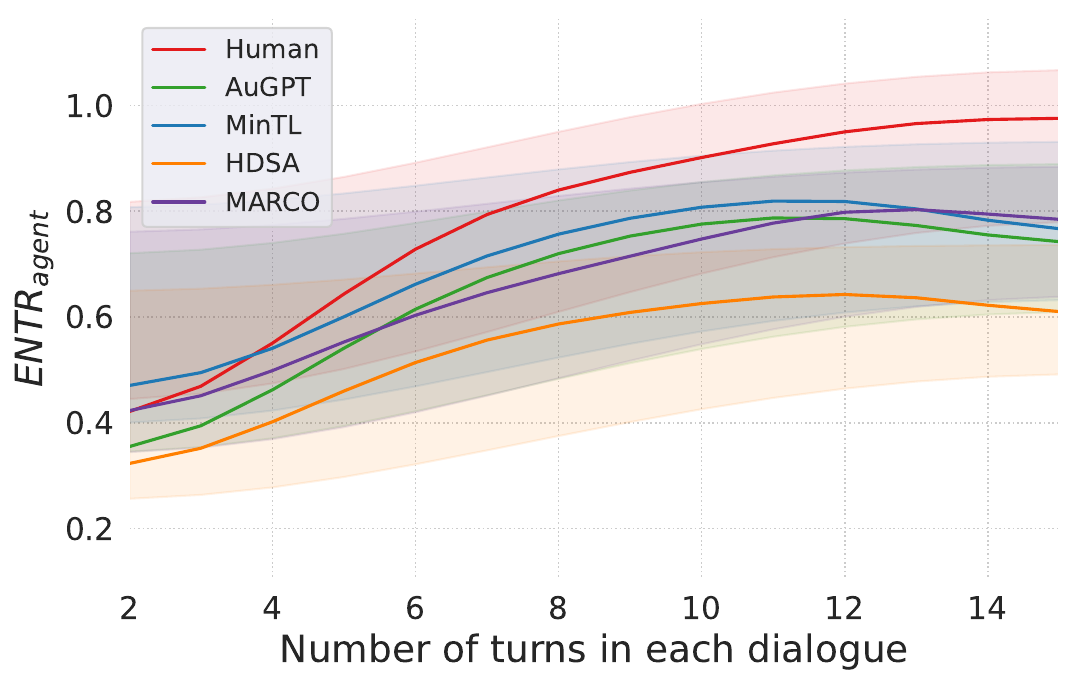}
  \caption{$\textsc{Entr}_{agent}$ wrt the number of turns. Solid lines represent mean values of $\textsc{Entr}_{agent}$ with a 1-D Gaussian filter. Shaded regions stand for the standard deviation.}
  \label{fig:entr_line_plot}
\end{figure}
\input{table/experiment1}

We also compute the $\textsc{Entr}_{agent}$ against the number of turns in dialogues. As shown in Figure~\ref{fig:entr_line_plot}, the degree of \lee tends higher as the number of turns increases, in line with expectations that more \lee happens when people speak more. The curves of response generation models are mainly under their counterparts for humans, indicating discrepancies between generated responses and human-level responses.

\subsection{\lee Extraction Task}
\label{sec:experiment}
We first compare the performance of two different baseline approaches and then evaluate the effect of different domains on the model performance.

\paragraph{\textbf{Experimental setup.}}
We conduct training for our models using three distinct input settings.
Firstly, as our primary objective is to model a human-like agent rather than replicate the speaking style of users, we exclusively train our models on agent samples. 
This yields $31\,436$, $4\,093$, and $4\,211$ samples for the training, validation, and testing sets.
Secondly, we investigate the impact of user samples by adding them to the training set, resulting in a total of $50\,130$ training samples. 
Thirdly, we consider the potential usefulness of incorporating dialogue act information, as dialogue acts can provide valuable information that may influence the next utterance.
Furthermore, Table \ref{tbl:statistics_wrt_expressions} reveals that the median and mode of the priming distance are both 1, indicating that speakers are more likely to use a \lee expression immediately after it appears.
Thus, we vary the dialogue history input range, including the last one, last two, or full history, denoted as ${1, 2, \text{full}}$.

\paragraph{\textbf{Comparison between end2end and pipeline approaches.}}
In Table~\ref{tbl:experiments1}, we observe that introducing user samples or agent acts into the training data does not improve the performance of the baseline models. This is consistent with our finding in Section \ref{sec:dataset_statistics} that there are few overlaps between \lee expressions and dialogue acts.
Moreover, we present a breakdown of the model's performance based on the range of dialogue contexts for each input setting.
Experimental results for the end-to-end approach reveal that decreasing the length of dialogue context has a negative impact on performance by significantly reducing the recall rate.
Although the model's attention is focused on a smaller set of dialogue contexts (precision rate increases), it is insufficient to offset the loss of missing some instances in the earlier dialogue context (recall rate decreases).
For the pipeline approach, the inclusion of false positives and the low recall rate at the first named entity steps lead to a poorer performance compared to the end-to-end approach.


\input{table/experiment2}
\paragraph{\textbf{The effect of different domains.}}
In Section \ref{sec:dataset_statistics}, we performed an ANOVA test which provided strong evidence of differences in the mean of lexicon sizes for the Taxi, Hotel, Attraction, Train, and Restaurant domains.
Thus, we proceeded to evaluate the proposed end-to-end model with respect to these five domains while excluding the hospital and police domains, which do not have a validation and testing set, following previous works on task-oriented conversation systems \cite{wu2019transferable,kim2020efficient,zhu2020efficient}. Furthermore, our analysis was restricted to agent samples only.

Table \ref{tbl:experiments2} presents the performance of the baseline model on the \multiwozentr dataset across the five domains. The experimental results indicate that the $F_1$ scores for the five domains fall within a narrow range, with no clear variation in the model's performance. However, training the model on a single domain resulted in lower performance compared to training it on all domains jointly, indicating that the model may benefit from more training data.

%% file: table/anaylsis.tex
\begin{table}[!ht]
\centering
\footnotesize
\caption{Average $ENTR_{agent}$ on the test set. $\dag$ indicates {\footnotesize $p<0.001$} statistical significance.}
\begin{adjustbox}{max width=\columnwidth}
\begin{tabular}{cccccc}
\toprule
\bf Human           & \bf \augpt        & \bf \mintl        & \bf \hdsa         & \bf \marco       \\ 
\midrule
0.798                  & $0.676^\dag$ & $0.558^\dag$ & $0.654^\dag$ & $0.719^\dag$ \\
\bottomrule
\end{tabular}
\end{adjustbox}
\label{tbl:analysis}
\end{table}


%% file: table/experiment1.tex
\newcommand\myfonts{\fontsize{8.3pt}{9.96pt}\selectfont}
\newcommand\myfontsize{\fontsize{8pt}{9.6pt}\selectfont}
\newcommand{\up}[1]{\myfontsize ($\textcolor{green}{\blacktriangle}#1\%)$}
\newcommand{\down}[1]{\myfontsize ($\textcolor{red}{\blacktriangledown}#1\%)$}

\begin{table*}[!ht]
    \centering
    \caption{
    The test results for the Lexical Entrainment Extraction Task under different settings, with colored arrows indicating the corresponding changes in performance relative to the baseline results. The best-performing results are highlighted in light orange and blue highlights the second place.
    }
    \begin{adjustbox}{max width=\textwidth}
    \begin{tabular}{lcccccccccc}
        \toprule
            & \multicolumn{3}{c}{\bf Dialogue History Length = 1} 
            & \multicolumn{3}{c}{\bf Dialogue History Length = 2}
            & \multicolumn{3}{c}{\bf Full Dialogue History Length}\\
            \cmidrule(lr){2-4}
            \cmidrule(lr){5-7}
            \cmidrule(lr){8-10}
            & $\text{\bf Recall}$ & $\text{\bf Precision}$ & $\text{\bf $\mathbf{F_1}$ Score}$ 
            & $\text{\bf Recall}$ & $\text{\bf Precision}$ & $\text{\bf $\mathbf{F_1}$ Score}$
            & $\text{\bf Recall}$ & $\text{\bf Precision}$ & $\text{\bf $\mathbf{F_1}$ Score}$ \\

        \midrule
        \rowcolor{Gray} \multicolumn{10}{l}{\textit{*End-to-End Approach}} \\
        Baseline        & 30.7           & 58.3                   & 40.3           & 40.5           & 49.7           & 44.7           & \se 70.0                 & \se  49.3            & \se 57.8 \\
        + user samples  & 33.9\up{10.4}  & 52.0\down{10.8}        & 41.0\up{1.7}   & 40.8\up{0.7}   & 47.9\down{3.6} & 44.1\down{1.3} &\blue 72.8\up{4}           &\blue 45.9\down{6.9}  &\blue 56.3\down{2.6} \\
        + agent acts    & 28.2\down{8.1} & 65.5\up{12.3}          & 39.4\down{2.2} & 38.7\down{4.4} & 54.6\up{9.9}   & 45.3\up{1.3}   & 77.7\up{11}          & 39.9\down{19.1} & 52.7\down{8.8} \\

        \midrule
        \rowcolor{Gray} \multicolumn{10}{l}{\textit{*Pipeline Approach}} \\
        Baseline     & 34.9         & 39.4             & 37.0            & 40.8         & 33.3            & 36.7            & 53.8         & 21.4           & 30.7 \\
        + user samples      & 35.5\up{1.7} & 40.6\up{3.0}     & 37.9\up{2.4}    & 41.5\up{1.7} & 35.1\up{5.4}    & 38.1\up{3.8}    & 53.5\up{0.6} & 22.7\up{0.6}   & 31.8\up{0.6}  \\
        + agent acts & 36.9\up{5.7} & 21.1\down{46.4}  & 26.9\down{27.3} & 44.6\up{9.3} & 12.2\down{63.4} & 19.1\down{48.0} & 55.7\up{4.1} & 7.8\down{53.6} & 13.7\down{55.4} \\
        \bottomrule
    \end{tabular}
    \end{adjustbox}
    \label{tbl:experiments1}
\end{table*}

%% file: table/experiment2.tex
\begin{table}[hbt!]
\centering
\footnotesize
\caption{Test results for five different domains using the end-to-end approach.
We train the model using the agents' utterances only.
The best performance for each column is highlighted in bold.
}
\begin{adjustbox}{max width=\columnwidth}
\begin{tabular}{lcccc}
\toprule
\bf Domains    & \bf Recall & \bf Precision & \bf $\mathbf{F_1}$ Score \\ \midrule
Taxi       & 63.53  & 44.17     & 52.11    \\
Hotel      & 68.79  & 42.16     & 52.28    \\
Attraction & 62.56  & 49.80     & 55.01    \\
Train      & 64.68  & 45.85     & 53.66    \\
Restaurant & 63.77  & 48.11     & 54.84    \\ 
\hdashline\noalign{\vskip 0.4ex}
All       & \bf 70.0    & \bf  49.3 & \bf  57.8 \\
\bottomrule
\end{tabular}
\end{adjustbox}
\label{tbl:experiments2}
\end{table}

%% file: paper/8_related_work.tex
\section{Related work}
\label{sec:related_work}

\paragraph{\textbf{Conversational systems}.}
Conversational AI has been a long-standing area of exploration in the natural language processing (NLP) research community \cite{10.1145/3209978.3210183,10.1145/3539813.3545126,hendriksen2020analyzing,stepGame2022shi,10.1145/3523227.3551481,ramos2022condita,shi2023dont,shi2023dept,10.1007/978-3-030-99736-6_20,rahmani-etal-2023-survey}. 
There are broadly two categories of conversational systems, open-domain \cite{wu2017sequential,huang2020challenges} and task-oriented \cite{hosseini2020simple,10.1145/3539618.3592041,hendriksen2021lstm,Sen_Wang_Xu_Yilmaz_2023}. Open-domain conversational systems mainly focus on open-ended conversations such as chit-chat, while task-oriented conversational systems target accomplishing a goal described by users in natural language such as hotel booking, movie recommendation, collaborative building \cite{budzianowski2018multiwoz,heck2020trippy,Shi2022learning,shi-etal-2023-rethinking,10.1145/3477495.3531718,10.1007/978-3-030-45439-5_14,veldkamp2023towards}. 
A number of corpora with various domains and annotations have been collected for task-oriented conversational systems. Some datasets \cite{henderson2014second,asri2017frames} contain conversations with a single domain, while others \cite{budzianowski2018multiwoz,byrne2019taskmaster,eric2019multiwoz} contain multi-domains conversations. 
\multiwoz dataset \cite{budzianowski2018multiwoz,eric2019multiwoz} spans 7 distinct domains with over $10\,000$ dialogues.

\paragraph{\textbf{Lexical Entrainment (\lee) for conversational systems}.} 
Some previous works \cite{hirschberg-2008-speaking,brockmann2005modelling,bakshi-etal-2021-structure,10.5555/1402383.1402416,campano2015like,duvsek2016context} attempted to integrate \lee in the conversationAL system.
\citet{10.5555/1402383.1402416} presented an entrainment model for aligning the language use of a virtual agent to the level of politeness and formality displayed by the user’s utterances.
 \citet{campano2015like} proposed an embodied conversational agent with the ability to choose when to share appreciation with a human partner in a museum setting.
 \citet{lopes2015rule} used rule-based and statistical models for the integration of \lee of an information-providing task in the public transport domain for spoken dialogue systems.
%
%
 \citet{duvsek2017novel} implemented a context-aware generator to model to entrain the user’s way of speaking by learning implicitly from data. 
 \citet{hu2016entrainment} designed a rule-based natural language generator to produce utterances entrained to a range of utterance features used in prior utterances by a human user, including referring expressions, syntactic template selection, and tense/modal choices.

%% file: paper/9_conclusion.tex
\section{Conclusion}
\label{sec:conclusion}

In this paper, we highlight the importance of incorporating \lee into conversational systems and propose a new dataset, named \multiwozentr, designed to facilitate the study of \lee. 
We conduct a comprehensive statistical analysis of the \multiwozentr dataset, providing valuable insights into its properties. 
Our investigation of current state-of-the-art response generation models highlights the issue of neglecting \lee, and our proposed measure serves as evidence for the lack of \lee in these models. 
To address this deficiency, we introduce two new tasks, namely, \lee extraction and generation, and propose two baseline models for the \lee extraction task under various settings. Our work lays the foundation for further research on the \lee generation task.

%% file: paper/9_limitation.tex
The discussion above is based on the assumption that researchers and designers should pursue the most human-like agents.
However, there is a risk of experiencing "an eerie sensation" \textemdash
\citet{mori2012uncanny} noted that some specific human-likeness designs may lead to an agent becoming repulsive, that is, falling into the uncanny valley. For example, people could be startled because of the prosthetic hand's limp boneless grip together with its texture and coldness, although the prosthetic hand has achieved a degree of resemblance to the human form.
In this paper, we are not arguing for the necessity to precisely mimic human beings. 
Rather we agree with \citet{amershi2019guidelines} that Artificial Intelligence (AI) systems should still follow social norms rather than “pretend” to be human beings. We believe, as \citet{thomas2021theories} suggested, that the study of natural phenomena in human-human conversations could give us inspiration for the design of machines, which is the goal of this work.

%% file: paper/10_appendix.tex
\section*{Appendix Overview}
The appendix is structured as follows:
\paragraph{Appendix \sect{sec:additional_measure}.} introduces several measurements for $\lee$.
\paragraph{Appendix \sect{sec:dataset_statistics}.} presents statistical analysis of the proposed \multiwozentr datasets.
\paragraph{Appendix \sect{sec:appendix_case_studies}.} provides three examples of the \multiwozentr dataset.
\paragraph{Appendix \sect{sec:filter_rules}.} provides examples of rule-based filters used in the dataset construction.
\paragraph{Appendix \sect{sec:training_details}.} provides implementation details and hyperparameters used in our experiments.

\section{Other Measures of \lee}
\label{sec:additional_measure}
In this section, we introduce several additional measures for \lee, adapted from \citet{nenkova2008high, duplessis2017automatic}:

\paragraph{\textbf{Expression Lexicon Size (ELS)}.} Let $\mathcal{E}$ be the set of all established \lee expressions given a dialogue. Then the expression lexicon size is the number of established \lee expressions given a dialogue, that is, $|\mathcal{E}|$, where $|\cdot|$ represents the cardinality of the set. The ELS is 10 for the dialogue in Table \ref{tbl:example}.

\paragraph{\textbf{Initial Expression Ratio ($\text{IER}_s$)}.} The initial expression ratio is the percentage of established \lee expressions initiated by the speaker $s$, and is computed as:
\begin{equation}
    IER_s = \frac{|\{e \in \mathcal{E}: e \text{ initiated by } s\}|}{|\mathcal{E}|},
\end{equation}
where $IER_{human}$ is 40\% and $IER_{agent}$ is 60\% in Table \ref{tbl:example}.

\paragraph{\textbf{Expression Repetition Ratio ($\text{ERR}_s$)}.} Let $\mathcal{T}_\mathcal{E}$ be the set of all tokens from all instances of established \lee expressions from a dialogue, and $\mathcal{T}$ be the set of all tokens from this dialogue, where $\mathcal{T}_\mathcal{E}$ is a subset of $\mathcal{T}$. 
Then the expression repetition ratio is the number of tokens from \lee expressions in the speaker $s$'s utterances divided by the total number of tokens in a dialogue:
\begin{equation}
    ERR_s = \frac{|\{t \in \mathcal{T}_\mathcal{E}: t \text{ from speaker } s\}|}{|\mathcal{T}|}
\end{equation}
For example, there are total $302$ tokens for the dialogue in Table \ref{tbl:example}. Among $302$ tokens, $17$ are from instances uttered by the user and $21$ are from instances uttered by the agent. Thus, in this case, $ERR_{user}$ is 5.6\% and $ERR_{agent}$ is 7.0\%.

\paragraph{\textbf{Frequency}.} Frequency refers to the number of utterances in which the established \lee expression appears. For example, in Table \ref{tbl:example}, the frequency of ``reference number'' is 3.

\paragraph{\textbf{Size}.} Size refers to the number of tokens which the established \lee expression is made up of. For example, in Table \ref{tbl:example}, the size of ``restaurant'' is 1, and the size of ``reference number'' is 2.

\paragraph{\textbf{Span}.} Span refers to the number of utterances in which the first appearance of the established \lee expression comes across to its last appearance, where both the first and last utterances count. For example, in Table \ref{tbl:example}, the span of ``reference number'' is 11, and the span of ``day'' is 2.

\paragraph{\textbf{Density}.} Density refers to the ratio between the established \lee expression's frequency and its span. For example, in Table \ref{tbl:example}, the density of ``ask'' is 0.8, and the density of ``price range'' is 0.3.

\paragraph{\textbf{Priming}.} Priming refers to the number of repetitions of the established \lee expression by the initiator before being used by the other interlocutor. In Table \ref{tbl:example}, the priming of ``price range'' is 2.

\paragraph{\textbf{Priming Distance}.} Priming Distance refers to the number of utterances in which the first appearance of the established \lee expression comes across to its first appearance from another speaker. For example, in Table \ref{tbl:example}, the priming distance of ``italian restaurant''.

\section{Statistical Analysis}
\label{sec:dataset_statistics}
This process results in the \multiwozentr dataset, consisting of $12\,038$, $1\,000$, and $1\,000$ dialogues for the training, validation, and testing sets with $62\,961$ \lee expressions. We present a thorough statistical analysis of the \multiwozentr dataset below.

\begin{figure}[!t]
  \centering
  \includegraphics[width=\columnwidth]{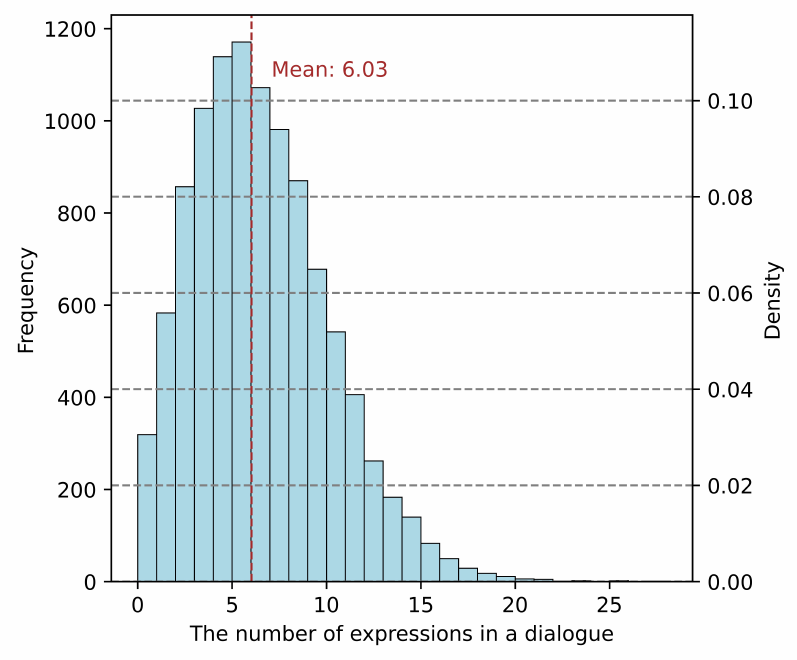}
  \caption{Distribution of Expression Lexicon Size (ELS).}
  \label{fig:distribution}
\end{figure}

\paragraph{\textbf{Measures' statistics}.} 
The first step of our statistical analysis is to compute the expression lexicon size (ELS) for each of the $10,438$ dialogues in the \multiwozentr dataset, as defined in \cref{sec:additional_measure}.
As depicted in Figure \ref{fig:distribution}, the ELS varies across the dialogues, with an average of 6.03.
Approximately $15\%$ of the dialogues have more than 10 \lee expressions.
In contrast, $288$ of $10\,438$ dialogues have zero \lee expression.
The maximum ELS value for a single dialogue is 30.

\input{table/statistics_speaker}

Table~\ref{tbl:statistics_wrt_speaker} presents the statistical analysis of two dialogue-level measures, initial expression ratio (IER) and expression repetition ratio (ERR).
The results indicate that the ERR values for the agent are higher than those for the human, suggesting that the agent's role players are more likely to entrain than the human's role players.
This finding aligns with our observation that about 75\% of the \lee expressions are initiated by the human speaker.
It is reasonable to assume that the agent's utterances are more likely to be entrained when the human speaker initiates more \lee expressions since the human speakers are those initiating the conversations and making requests. 
As the \multiwoz dataset involves human-to-human interactions, this implies that users are more likely to initiate a \lee expression, and a human-like agent's responses need to incorporate a higher degree of \lee to make conversations more engaging and successful. 
This underscores the importance of integrating \lee expressions into agent utterances to enhance their effectiveness.

\input{table/statistics_expression_level}
The statistical analysis of expression-level measures is presented in Table \ref{tbl:statistics_wrt_expressions}.
The median and mode of the priming distance are both 1, which means that a large proportion of the \lee expressions (63.79\%) occur within two turns of the conversation.
This finding suggests that the majority of \lee expressions are produced immediately after the first use of the expression.
To investigate the effects of this phenomenon on the model's performance, we limit the model's attention to the last two utterances of the preceding dialogue context in the entrainment extraction task in \cref{sec:experiment}.


\paragraph{\textbf{The effect of different domains}.} 
To investigate the penitential effects of different domains on \lee, we compute the expression lexicon size with respect to different domains.
We observe that the Police and Hospital domains have relatively smaller lexicon sizes compared to the other five domains. However, it should be noted that these two domains have fewer training samples and no validation or testing samples.
To determine if there are statistically significant differences between the domains, we conducted a one-way ANOVA test with the null hypothesis $H_0$: $\mu_{Taxi} = \mu_{Hotel} = \mu_{Attraction} = \mu_{Train} = \mu_{Restaurant}$, where $u_d$ represents the mean of the expression lexicon size for domain $d$.
Our experimental results indicate that the null hypothesis can be rejected with a p-value of $2.59\mathrm{e}^{-43}$, suggesting strong evidence of differences in the mean of lexicon sizes among the five domains with large sample sizes. 
Therefore, we further examine the effects of different domains on the entrainment extraction task in Section \ref{sec:experiment}.



\paragraph{\textbf{The effect of dialogue acts}.} 
In the response generation process, the entrainment extraction module's output is used in conjunction with the outputs from other modules, such as dialogue state tracking and intent detection, as shown in Figure \ref{fig:flow_chart}.
It is essential to understand the extent of overlap among these outputs. 
To address this issue, we examine the lexicon overlap between \lee expressions and dialogue acts.
We stem each token with the NLTK toolkit and obtain $17\,594$ unique tokens for dialogue acts and $3\,477$ for \lee expressions, with only 874 tokens belonging to both sets.
The analysis shows that the lexicon of \lee expressions shares only a small fraction of overlap with the lexicon of dialogue acts, indicating that the degree of overlap with original forms and contextual meaning is also likely to be low. This suggests that the incorporation of dialogue acts may have a limited effect on the \lee extraction task. We evaluate the impact of incorporating dialogue acts on the entrainment extraction task empirically in \cref{sec:experiment}.

\section{Case Studies}
\label{sec:appendix_case_studies}
In Table~\ref{tbl:case_study_1_appendix},~\ref{tbl:case_study_2_appendix} and~\ref{tbl:case_study_3_appendix}, we provide three examples as a supplementary of~\cref{sec:model_inspection}.

\section{Rules Used in Dataset Construction}
\label{sec:filter_rules}
We provide two examples of rules defined in~\cref{sec:dataset} to check whether "help" and "booking" are verbs.
For example, we find all cases where "booking" is used as a verb. We add one space before "in booking" to avoid "booking" in "train booking" being counted as a verb. There are many similar words such as "contact" and "cost".
For the sake of the space, we will not list all these rules. The code will be released after this paper is accepted.

\definecolor{codegreen}{rgb}{0,0.6,0}
\definecolor{codegray}{rgb}{0.5,0.5,0.5}
\definecolor{codepurple}{rgb}{0.58,0,0.82}
\definecolor{backcolour}{rgb}{0.95,0.95,0.92}

\lstdefinestyle{mystyle}{
    backgroundcolor=\color{backcolour},   
    commentstyle=\color{codegreen},
    keywordstyle=\color{magenta},
    numberstyle=\tiny\color{codegray},
    stringstyle=\color{codepurple},
    basicstyle=\ttfamily\footnotesize,
    breakatwhitespace=false,         
    breaklines=true,                 
    captionpos=b,                    
    keepspaces=true,                 
    numbers=left,                    
    numbersep=5pt,                  
    showspaces=false,                
    showstringspaces=false,
    showtabs=false,                  
    tabsize=2
}

\lstset{style=mystyle}

\begin{lstlisting}[language=Python]{BitXorMatrix.m}
def special_case_booking(utterance: str) -> bool:
    """
    Check whether booking is a verb given an utterance
    """
    special_cases = [
        'are you booking', ' before booking', 
        'booking your ', ' i booking',
        'booking the ', 'booking that',
        ' in booking', ' be booking'
    ]
    for sc in special_cases:
        if sc in utterance:
            return True
    return False

def special_case_help(utterance: str) -> bool:
    """
    Check whether help is a verb given an utterance
    """
    special_cases = [
        'can you help', "can I help", 
        'could you help', 'help you',
        "could help", 'would help',
        'may help', 'might help',
        'help me', 'will help', 
        "to help", "can help", 
        "can you please help", 
    ]
    for sc in special_cases:
        if sc in utterance:
            return True
    return False
\end{lstlisting}

\input{table/appendx_hyparam.tex}

\input{table/case_study_appendix}

\section{Training Details}
\label{sec:training_details}
During the tokenization process of \textsc{Bert}, any previously unseen words are segmented into sub-words, which results in the creation of multiple tokens for a single word. To obtain a score for the entire word, we use the score assigned to the first sub-word and exclude the remaining sub-words when calculating the cross-entropy loss.
Additionally, we exclude the special tokens \texttt{[CLS]}, \texttt{[SEP]}, and \texttt{[PAD]} when calculating the loss.
In scenarios where the dialogue history is incomplete, some instances of \lee expressions may not be detected. To ensure unbiased evaluation, we consider instances of these expressions outside the dialogue history as false negative examples. 
When the length of the dialogue history is not complete, it is inevitable that certain instances of \lee expressions will not be captured. To account for this, instances of these expressions outside the dialogue history are automatically counted as false negative examples to ensure a fair evaluation when compared to scenarios where the dialogue history is complete.
To evaluate the end-to-end approach, we assess its performance at the noun phrase level, calculating the recall rate, precision rate, and $F_1$ score. The positive weight $\alpha$ for the cross-entropy loss is varied between 4 and 12.

%% file: table/statistics_speaker.tex
\begin{table}[htbp]
\centering
\caption{Statistics with respect to speakers: $a$ represents agent and $u$ represents human.}
\resizebox{0.8\columnwidth}{!}{
\begin{tabular}{lcccc}
\toprule
         & $\bf IER_{u}$  & $\bf IER_{a}$  & $\bf ERR_{u}$ & $\bf ERR_{a}$  \\ 
\midrule
\bf Mean.                & 75.02      & 24.98      & 5.29      & 5.76       \\
\bf Std.                 & 21.39      & 21.39      & 2.59      & 2.94       \\ 
\bottomrule
\end{tabular}
}
\label{tbl:statistics_wrt_speaker}
\end{table}

%% file: table/statistics_expression_level.tex
\begin{table}[!ht]
\centering
\caption{Statistics with respect to $62\,961$ expressions: Distance stands for the priming distance.}
\begin{adjustbox}{max width=\columnwidth}
\begin{tabular}{lcccccc}
\toprule
 & \bf Mean.  & \bf Std. & \bf Min. &  \bf Max.  & \bf Median.  & \bf Mode. \\ \midrule
\bf Frequency & 2.62  & 0.98          & 2          &  17 & 2   & 2 \\
\bf Size      & 1.47  & 0.88         & 1          &  17 & 1    & 1 \\
\bf Span      & 5.02  & 4.05     & 2         &  40 & 4    & 2 \\
\bf Priming   & 1.11  & 0.34  & 1 &  6  & 1     & 1 \\
\bf Distance  & 2.55  & 2.90  & 1 &  35 & 1     & 1 \\
\bf Density   & 0.72  & 0.30  & 0.06  &  2.00  & 0.75    & 1.00 \\
\bottomrule
\end{tabular}
\end{adjustbox}
\label{tbl:statistics_wrt_expressions}
\end{table}

%% file: table/appendx_hyparam.tex
\begin{table}[!t]
    \centering
    \tiny
    \caption{Hyperparameters for the \lee extraction task.}
    \resizebox{0.9\columnwidth}{!}{
    \begin{tabular}{cc}
        \toprule
        \textbf{Hyperparameter} & \textbf{Assignment}  \\
        \midrule
        number of steps & 100 epochs or 50K steps\\
        \midrule
        batch size & 24\\
        \midrule
        maximum learning rate & 5e-5 \\
        \midrule
        learning rate optimizer & AdamW \\
        \midrule
        max sequence length & 512 \\
        \midrule
        Adam epsilon & 1e-6 \\
        \midrule
        Adam beta weights & 0.9, 0.98\\
        \midrule
        learning rate scheduler & Warmup linear \\
        \midrule
        Weight decay & 0.01 \\
        \midrule
        Warmup proportion & 0.06 \\
        \midrule
        learning rate decay & linear \\
        \midrule
        fusion layers & 2 \\
        \bottomrule
    \end{tabular}}
    \label{table:hyperparameters}
\end{table}

%% file: table/case_study_appendix.tex
\begin{table*}[hbt!]
    \centering
    \footnotesize
    \setlength{\tabcolsep}{5.75pt}
    \renewcommand{\arraystretch}{1.2}
    \caption{Example with two agent responses. Current response generation models could partially entrain the user's way of speaking such as "\textit{contact number}". However, it is not optimal for these generation models to use "\textit{taxi}" when users are using "\textit{cab}".}
    \resizebox{\textwidth}{!}{
    \begin{tabular}{p{0.1\linewidth}|p{0.88\linewidth}}
      \toprule
      \bf Model & \bf Utterance \\
      \midrule
      \textsc{User}  &  I need a \colorbox{color5}{cab} too please, arriving by my \colorbox{color4}{booked time}, car type and \colorbox{brown!45}{contact number} would be helpful too please. \\
      Agent & You need a \colorbox{color5}{\underline{cab}} to get you to the restaurant before the \colorbox{color4}{\underline{booked time}} and where do you need \colorbox{color5}{\underline{cab}} to pick you up at? \\
      \textsc{User}  & From the Cambridge Punter, please. \\
      \textsc{Agent} & A black tesla is booked, with \colorbox{brown!45}{\underline{contact number}} 07774975314.  \\
      \midrule
      \multicolumn{2}{l}{\bf Generated responses for the first agent response} \\
      \augpt &  I have booked a blue skoda for you. The contact number is 85534112262. \\
      \mintl &  I have booked you a BMW . The contact number is 07338019809 . Where will you be departing from ? \\
      \hdsa  &  Where will you be departing from and where are you going ? \\
      \marco &  I would be happy to book a \colorbox{color10!70}{\underline{taxi}} for you . where are you departing from and what is your destination ? \\
      \midrule
      \multicolumn{2}{l}{\bf Generated responses for the second agent response} \\
      \augpt &  I have booked you a black honda and the \colorbox{brown!45}{\underline{contact number}} is 91788856987. Is there anything else I can help you with? \\
      \mintl &  I have booked your taxi . Be expecting a BMW . If you need to reach them , please call 07338019809 . \\
      \hdsa  &  I have booked you a \texttt{[taxi\_type]} . the \colorbox{brown!45}{\underline{contact number}} is \texttt{[taxi\_phone]} . is there anything else i can help you with ? \\
      \marco &  I have booked a \texttt{[taxi\_type]} for you . the \colorbox{brown!45}{\underline{contact number}} is  \texttt{[taxi\_phone]} . is there anything else i can help you with ? \\
      \bottomrule
    \end{tabular}
    }
    \label{tbl:case_study_1_appendix}
\end{table*}
\begin{table*}[hbt!]
    \centering
    \footnotesize
    \setlength{\tabcolsep}{5.75pt}
    \renewcommand{\arraystretch}{1.2}
    \caption{Example with one agent response. In the ground-truth agent response, lexical entrainment with the user's requirement including "\textit{Monday}" and "\textit{5 nights}" makes the response more informative and less ambiguous.}
    \resizebox{\textwidth}{!}{
    \begin{tabular}{p{0.1\linewidth}|p{0.88\linewidth}}
      \toprule
      \bf Model & \bf Utterance \\
      \midrule
      \textsc{User}  & Yes, I would like to book a stay for 1 person, starting \colorbox{color4}{Monday} and for \colorbox{color5}{5 nights}. \\
      \textsc{Agent} & Your room has been booked at the Bridge Guest House for \colorbox{color5}{\underline{5 nights}} starting \colorbox{color4}{\underline{Monday}}. Your reference code is 7VM9FIPC. Is there anything else I can do to help you?  \\
      \midrule
      \multicolumn{2}{l}{\bf Generated responses for the first agent response} \\
      \augpt &  I have made those reservations and your reference number is~\colorbox{white}{00000011.} \\
      \mintl &  I have booked your room. Your reference number is~\colorbox{white}{QJTIXK8C.} \\
      \hdsa  &  Booking was successful. Reference number is 7vm9fipc. Is there anything else i can help you~\colorbox{white}{with?} \\
      \marco &  I was able to book you at the \texttt{[hotel\_name]} for \texttt{[value\_count]} nights. Your reference number is 7vm9fipc. Is there anything else i can help you with? \\
      \bottomrule
    \end{tabular}
    }
    \label{tbl:case_study_2_appendix}
\end{table*}
\begin{table*}[hbt!]
    \centering
    \footnotesize
    \setlength{\tabcolsep}{5.75pt}
    \renewcommand{\arraystretch}{1.2}
    \caption{Example with one agent response. In the ground-truth agent response, lexical entrainment with the user's requirement including "\textit{8 people}" and "\textit{4 nights}" makes the response more informative and less ambiguous.}
    \resizebox{\textwidth}{!}{
    \begin{tabular}{p{0.1\linewidth}|p{0.88\linewidth}}
      \toprule
      \bf Model & \bf Utterance \\
      \midrule
      \textsc{User}  & I need it for \colorbox{color5}{8 people}, and for \colorbox{color10!70}{4 nights} starting Saturday. \\
      \textsc{Agent} & Great, I have booked your stay for \colorbox{color5}{\underline{8 people}} for \colorbox{color10!70}{\underline{4 nights}}. Your reference number is PB41WUOT. May I help with anything else? \\
      \midrule
      \multicolumn{2}{l}{\bf Generated responses for the first agent response} \\
      \augpt &  Booking was successful. Reference number is: 00000003. \\
      \mintl &  I have booked your room at Allenbell for saturday for \colorbox{color10!70}{\underline{4 nights}} for \colorbox{color5}{\underline{8 people}}. Your reference number is 6KNJSQXY. Is there anything else i can help you with? \\
      \hdsa  &  Booking was successful. Reference number is pb41wuot. Is there anything else i can help you with? \\
      \marco &  I was able to book you at the \texttt{[hotel\_name]} for \texttt{[value\_count]} nights. your reference number is pb41wuot. Is there anything else i can help you with? \\
      \bottomrule
    \end{tabular}
    }
    \label{tbl:case_study_3_appendix}
\end{table*}